\documentclass[runningheads]{llncs}

\usepackage{eccv}

\usepackage{eccvabbrv}

\usepackage{graphicx}
\usepackage{booktabs}

\usepackage[accsupp]{axessibility}  % Improves PDF readability for those with disabilities.

\usepackage{orcidlink}

\usepackage{multirow}
\usepackage{tcolorbox}
\usepackage{colortbl}      
\usepackage{wrapfig}
\usepackage{caption}
\usepackage{enumitem}
\usepackage{makecell}
\usepackage{marvosym}
\newcommand{\blfootnote}[1]{%
  \begingroup
  \renewcommand\thefootnote{}\footnote{#1}%
  \addtocounter{footnote}{-1}%
  \endgroup
}

\begin{document}

% ---------------------------------------------------------------
% TODO REVIEW: Replace with your title
\title{PhyMAGIC: Physical Motion-Aware Generative Inference with Confidence-guided VLM} 
% \title{PhyMAGIC: Physical Motion-Aware Generative Inference with Confidence-guided VLM} 

% TODO REVIEW: If the paper title is too long for the running head, you can set
% an abbreviated paper title here. If not, comment out.
\titlerunning{PhyMAGIC}

% TODO FINAL: Replace with your author list. 
% Include the authors' OCRID for the camera-ready version, if at all possible.
\author{Siwei Meng\inst{1}\orcidlink{0009-0008-3044-2842} \and
Yawei Luo\inst{2}\textsuperscript{\Letter}\orcidlink{0000-0002-7037-1806} \and
Ping Liu\inst{1}\textsuperscript{\Letter}\orcidlink{0000-0002-3170-3783}}

% TODO FINAL: Replace with an abbreviated list of authors.
\authorrunning{S.~Meng et al.}
% First names are abbreviated in the running head.
% If there are more than two authors, 'et al.' is used.

% TODO FINAL: Replace with your institution list.
\institute{Department of Computer Science \& Engineering, University of Nevada, Reno, USA \and
School of Software Technology, Zhejiang University, China \\
% \url{http://www.springer.com/gp/computer-science/lncs} \\
\email{siweim@unr.edu, yaweiluo@zju.edu.cn, pino.pingliu@gmail.com}}

\maketitle

\blfootnote{\textsuperscript{\Letter} Corresponding authors.}

\begin{abstract}
Inferring physical properties from a single image is fundamentally under-constrained.
Attributes such as density, elasticity, and yield stress govern how objects move, yet they are largely invisible in a static frame.
Existing physics-aware methods attempt to resolve this ambiguity through task-specific fine-tuning or supervised property estimation, but both strategies struggle to generalize across diverse materials and scenes.
We observe that different motions expose complementary physical cues.
Building on this observation, we propose \textbf{PhyMAGIC}, a training-free framework that actively probes physical properties by synthesizing targeted motions from a single image.
Specifically, PhyMAGIC uses a pretrained image-to-video model to construct \emph{motion probes} that generate diverse dynamic sequences from the input image.
A vision-language model then analyzes these sequences to estimate physical parameters, each accompanied by a confidence score.
Parameters with low confidence trigger targeted prompt refinement, which generates additional probe motions to gather complementary evidence.
Once all parameters reach sufficient confidence, PhyMAGIC compiles them into a complete physical specification and executes it in a differentiable Material Point Method simulator initialized from 3D Gaussian reconstructions.
Experiments on diverse real-world scenes demonstrate that PhyMAGIC achieves stronger text-motion alignment and higher human-rated physical plausibility than state-of-the-art open-source video generators and physics-aware baselines.
Code is available at: \url{https://mengsiwei.github.io/MAGIC/}.
\end{abstract}

\section{Introduction}
\label{sec:intro}

Recent advances in geometry and texture synthesis~\cite{xiang2024trellis_arxiv2024, voleti2024sv3d_eccv2024, chen2024text_cvpr2024, tang2024lgm_eccv2024, miao2025advances_arxiv2025} have significantly improved the fidelity of static 3D assets from a single image.
Extending these assets to physically consistent dynamics, however, remains an open challenge.
Modern video diffusion models~\cite{yang2024cogvideox_iclr2025, peng2025open_arxiv_2025} can produce visually realistic motion, but they lack explicit control over physical attributes such as mass, elasticity, and friction.
As a result, their outputs frequently exhibit momentum violations, object interpenetration, and unrealistic material responses.

These limitations have motivated research on physics-aware generative methods~\cite{meng2025grounding_arxiv2025, lin2025exploring_arxiv2025}.
Existing approaches generally follow two paradigms.
The first embeds physical priors directly into generative architectures~\cite{xu2024precise_arxiv2024, cao2024neuma_neurips2024}, which improves physical consistency but often reduces generalization to novel scenes or unseen materials.
The second relies on vision-based refinement or learning-based property estimation~\cite{liu2024physgen_eccv2024, tan2024physmotion_arxiv2024, liu2025physflow_cvpr2025}, which offers more flexibility but depends on large annotated datasets and remains vulnerable to perceptual ambiguity.
Both paradigms share a common bottleneck: they struggle to reliably infer intrinsic physical properties when visual evidence is scarce.

This bottleneck reflects a deeper issue: physical inference from a single static image is fundamentally under-constrained.
Multiple motion trajectories can appear visually plausible for the same object while being consistent with very different underlying physical properties.
Attributes such as density, elasticity, and yield stress are largely invisible in a static frame, yet they critically determine how an object moves, deforms, and interacts.
Manually specified priors~\cite{xie2024physgaussian_cvpr2024} lack adaptability across materials.
Learning-based estimators~\cite{cai2024gic_arxiv2024, zhang2024physdreamer_eccv2024, lin2025omniphysgs_iclr2025} are susceptible to data scarcity and domain shift.
Recent Vision-Language Models (VLMs)-based methods~\cite{liu2024physgen_eccv2024, liu2025physflow_cvpr2025} offer a promising direction by incorporating high-level physical reasoning, but they are less reliable when visual cues are sparse and are not coupled with generative models in a closed-loop fashion.
What is missing is a mechanism that can actively gather physical evidence from a single image, rather than passively predicting properties from limited observations.

\begin{figure}[t]
    \captionsetup{font=small}
    \centering
    \includegraphics[width=1.0\linewidth]{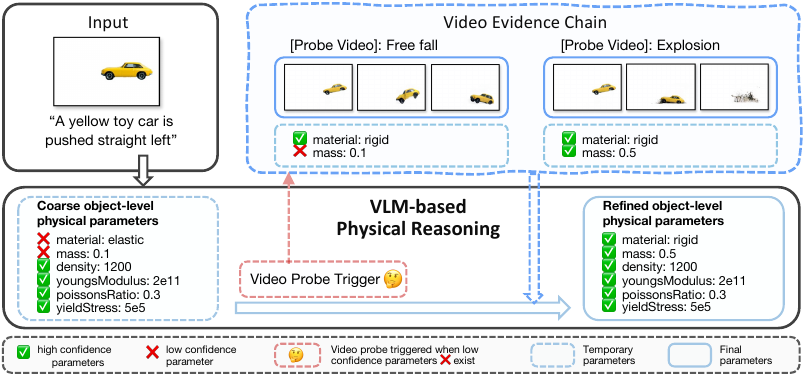}
    \caption{
  \textbf{PhyMAGIC: Active motion probing for physical identification.}
    A single image and instruction are under-constrained for inferring motion-critical physical attributes.
    For example, coarse VLM reasoning on the static input correctly identifies the material as rigid but assigns low confidence to mass and yield stress.
    To resolve this uncertainty, PhyMAGIC synthesizes targeted motion probes that expose complementary physical cues.
    The VLM re-evaluates the probed videos across iterations, progressively refining low-confidence parameters until a complete set of physical attributes is obtained.
    The resulting parameters, together with simulator-native descriptors are then executed in an MPM simulator to produce physically consistent dynamics.
    }
    \label{fig:motivation}
\end{figure}

% To address this challenge, we propose \textbf{PhyMAGIC}, a training-free framework that actively probes physical properties by synthesizing targeted motions from a single image.
% Our key insight is that pretrained image-to-video (I2V) diffusion models can serve as \emph{motion probes} that expose informative physical evidence for reliable VLM reasoning.
% As illustrated in \autoref{fig:motivation}, different motion patterns reveal complementary aspects of an object's physical properties: impact and free-fall motions expose inertia and stiffness cues that are difficult to observe from gentle perturbations alone.
% This formulation transforms physical inference from passive single-image prediction into an active probing process, where each generated motion serves as a targeted experiment to reduce uncertainty.

To address this challenge, we propose \textbf{PhyMAGIC}, a training-free framework that resolves the ambiguity of single-image physical reasoning through an active video evidence chain and generates physically controlled dynamics with a differentiable MPM simulator.
Instead of directly estimating physical attributes from a single static observation, PhyMAGIC synthesizes motion probes that expose otherwise hidden physical cues, which provides temporal evidence about object behavior.
Our key insight is that combining the reasoning ability of vision-language models (VLMs) with the motion diversity of pretrained image-to-video (I2V) diffusion models can actively reveal informative physical behaviors and progressively reduce uncertainty in physical inference.
As illustrated in~\autoref{fig:motivation}, different motion patterns reveal complementary aspects of an object's physical properties: free-fall and explosion motions expose material and mass cues that are difficult to observe from gentle perturbations alone.
This formulation transforms physical inference from passive single-image prediction into an active probing process, where each generated motion serves as a targeted experiment to reduce uncertainty.

Concretely, PhyMAGIC operates in three tightly coupled stages.
In the \emph{Motion Probe Generation stage}, PhyMAGIC synthesizes motion probe videos using a pretrained I2V model~\cite{yang2024cogvideox_iclr2025}, transforming a single frame into a rich source of temporal evidence.
In the \emph{VLM-based Physical Reasoning stage}, a VLM then analyzes these videos to estimate physical parameters, each accompanied by a confidence score.
Parameters with low confidence trigger targeted prompt refinement, which generates additional probe motions to gather complementary evidence.
This loop repeats until all parameters reach sufficient confidence or an iteration budget is exhausted.
This closed-loop design enables PhyMAGIC to iteratively reduce uncertainty in physical inference without manual intervention.
In the \emph{physics-grounded dynamic stage}, the VLM additionally infers simulator-native descriptors (e.g., boundary conditions, contact modes, and force directions) from the probe context and instruction.
PhyMAGIC compiles these descriptors together with the inferred material parameters into a hybrid physical specification.
This specification is executed in a differentiable Material Point Method (MPM) simulator~\cite{jiang2015affine_tog2015}, initialized from 3D Gaussian reconstructions~\cite{xiang2024trellis_arxiv2024}, producing physically consistent dynamic 3D content without task-specific training or manual annotation.

Our main contributions are as follows:
\begin{itemize}[leftmargin=*,topsep=0pt,itemsep=2pt,parsep=0pt]

    \item \textbf{Active motion probing for physical inference.}
    We formulate single-image physical reasoning as an active evidence acquisition problem.
    Pretrained I2V models serve as motion probes, and iterative VLM feedback progressively reveals complementary physical cues for motion-critical attributes.

    \item \textbf{Confidence-guided prompt refinement.}
    We introduce an iterative refinement strategy that uses per-attribute confidence scores to identify uncertain parameters and automatically generates targeted probe motions that reduce uncertainty.

    \item \textbf{Hybrid physical specification with differentiable simulation.}
    We introduce a unified representation that combines VLM-inferred material properties (density, elasticity, yield stress) with simulator-native descriptors (boundary conditions, contact modes, force configurations), and execute it in a differentiable MPM simulator for physics-consistent dynamic 3D generation.

    \item \textbf{Comprehensive evaluation.}
    Experiments on PhysGaussian, PhysGen, and Internet-collected scenes demonstrate that PhyMAGIC achieves stronger text-motion alignment and higher human-rated physical plausibility than state-of-the-art video generators and physics-aware baselines, while maintaining competitive visual quality.

\end{itemize}

\section{Related Work}
\label{sec:relatedworks}

\textbf{3D Dynamic Generation.}
Recent methods extend static 3D representations into temporal sequences via trajectory modeling or deformation fields~\cite{li2024dreammesh4d_neurips2024, ren2024l4gm_neurips2024}.
Neural implicit approaches~\cite{du2021nerfflow_iccv2021, pumarola2021d_cvpr2021} achieve photorealistic dynamic novel-view synthesis but at high computational cost.
3D Gaussian Splatting (3DGS)~\cite{kerbl3DGS_tog2023} and its dynamic extensions~\cite{katsumata2024compact_eccv2024,zhang2024dynamics} enable real-time rendering with limited deformations, yet these approaches focus on visual fidelity rather than explicit physical modeling.
As a result, the recovered dynamics may look plausible in pixel space but are not directly tied to material properties, forces, or boundary conditions.
PhyMAGIC builds on 3DGS but differs in that it explicitly injects VLM-inferred physical attributes into a differentiable simulator.
This turns 3DGS from a mainly appearance-driven representation into a simulation-ready physical carrier for downstream motion prediction and rendering.

\noindent\textbf{Physics-Grounded Generative Models.}
Physically embedded networks~\cite{xu2024precise_arxiv2024, cao2024neuma_neurips2024} integrate constraints such as elasticity and collision responses into architectures, but typically involve material-dependent designs that restrict generalization.
PhysGaussian~\cite{xie2024physgaussian_cvpr2024} simulates deformation via continuum mechanics with manually specified parameters.
PhysDreamer~\cite{zhang2024physdreamer_eccv2024} learns physical properties from video diffusion models, and Physics3D~\cite{liu2024physics3d_arxiv2024} combines viscoelastic MPM with Score Distillation Sampling for multi-material simulation.
OmniPhysGS~\cite{lin2025omniphysgs_iclr2025} learns constitutive relations to model diverse materials, but its generalization depends on training data coverage.
Vision-based refinement frameworks~\cite{liu2024physgen_eccv2024, liu2025physflow_cvpr2025, tan2024physmotion_arxiv2024} decouple generation from physical reasoning, offering flexibility but facing difficulties in precisely controlling physical attributes.
In contrast, PhyMAGIC infers parameters at test time through active motion probing, requiring no task-specific training or manual tuning.

\noindent\textbf{Vision-Language Model Reasoning for Physical Inference.}
VLMs have shown promise for physical reasoning by leveraging implicit knowledge of materials and motion~\cite{achiam2023gpt_arxiv2023,wang2025monosr_arxiv2025}.
PhysGen~\cite{liu2024physgen_eccv2024} infers mass, elasticity, and force directions from static images via a VLM, while PhysFlow~\cite{liu2025physflow_cvpr2025} enriches motion context through auxiliary video generation~\cite{yang2024cogvideox_iclr2025} before property estimation.
However, both perform inference in a single forward pass without mechanisms to refine uncertain predictions.
Learning-based estimators such as GIC~\cite{cai2024gic_arxiv2024} train dedicated networks but are susceptible to data scarcity and domain shift.
PhyMAGIC addresses these limitations through an iterative confidence-guided loop that actively generates targeted motion probes to gather complementary evidence for low-confidence attributes.

\section{Preliminary}
\label{sec:preliminary}
In this section, we introduce the mathematical foundations and notation underlying our framework, with a focus on Gaussian Splatting for 3D representation and the Material Point Method for differentiable physical simulation in 3D space.

\noindent\textbf{3D Gaussian Splatting (3DGS).}
3D Gaussian Splatting~\cite{kerbl3DGS_tog2023} represents a scene as a collection of anisotropic Gaussian kernels $\{x_k, \Sigma_k, \alpha_k, c_k\}, {k\in \mathcal K}$, encoding position, covariance matrix, opacity, and view-dependent color.
Unlike implicit neural representations~\cite{mildenhall2020nerf_eccv2020, zhang2021learning_iccv2021}, 3DGS features a fully differentiable rasterization rendering process, where the pixel color $C(u)$ is computed by alpha blending depth-sorted Gaussians:
% and the pixel color $C(u)$ is computed via alpha blending of depth-sorted Gaussians:
\begin{equation}
C(u) = \sum_{k \in \mathcal{P}(u)} \alpha_k \, \mathcal{G}_k(u) \, \mathrm{SH}(d_k; c_k) \prod_{j < k} \left(1 - \alpha_j \, \mathcal{G}_j(u)\right).
\end{equation}
Here $\mathcal{G}_k(u)$ is the current Gaussian projection at pixel $u$, and $\mathrm{SH}$ denotes spherical harmonics for view-dependent colors $c_k$ of viewing direction $d_k$. 
Crucially, 3DGS exposes per-Gaussian parameters that are directly optimizable, which we later exploit to reconstruct and simulate deformable objects from the single image input.
% Crucially, 3DGS provides directly optimizable properties where every Gaussian kernel can be individually modified to achieve efficient deformations.

\noindent\textbf{Continuum Mechanics and Material Point Method.}
We model material behaviors using continuum mechanics~\cite{jiang2015affine_tog2015,reddy2013introduction_2013} that describes local deformation by a deformation map $\phi(\mathbf{X},t)$, relating initial coordinates $\mathbf{X}$ to current positions $\mathbf{x}=\phi(\mathbf{X},t)$. 
For numerical simulation, we employ the Material Point Method (MPM)~\cite{jiang2015affine_tog2015,zong2023neural_siggraph2023}, a hybrid Lagrangian–Eulerian approach that discretizes materials into particles $\{\mathbf{x}_p, m_p, \mathbf{v}_p, \mathbf{F}_p\}_{p=1}^N$ carrying mass, velocity, and deformation state, while using a background grid to compute forces and spatial derivatives.
In MPM, each simulation step alternates between: (1) particle-to-grid (P2G) transfer of mass and momentum, (2) grid update by solving discretized momentum equations under internal and external forces, and (3) grid-to-particle (G2P) transfer to update particle velocities, positions, and deformation gradients.
This formulation naturally handles large deformations, collisions, and multi-material interactions with numerical stability, and its differentiable implementation allows seamless integration into our closed-loop generation framework.

\section{Methodology}
\label{sec:methodology}
\begin{figure}[t]
    \captionsetup{font=small}
    \centering
    \includegraphics[width=1.0\textwidth]{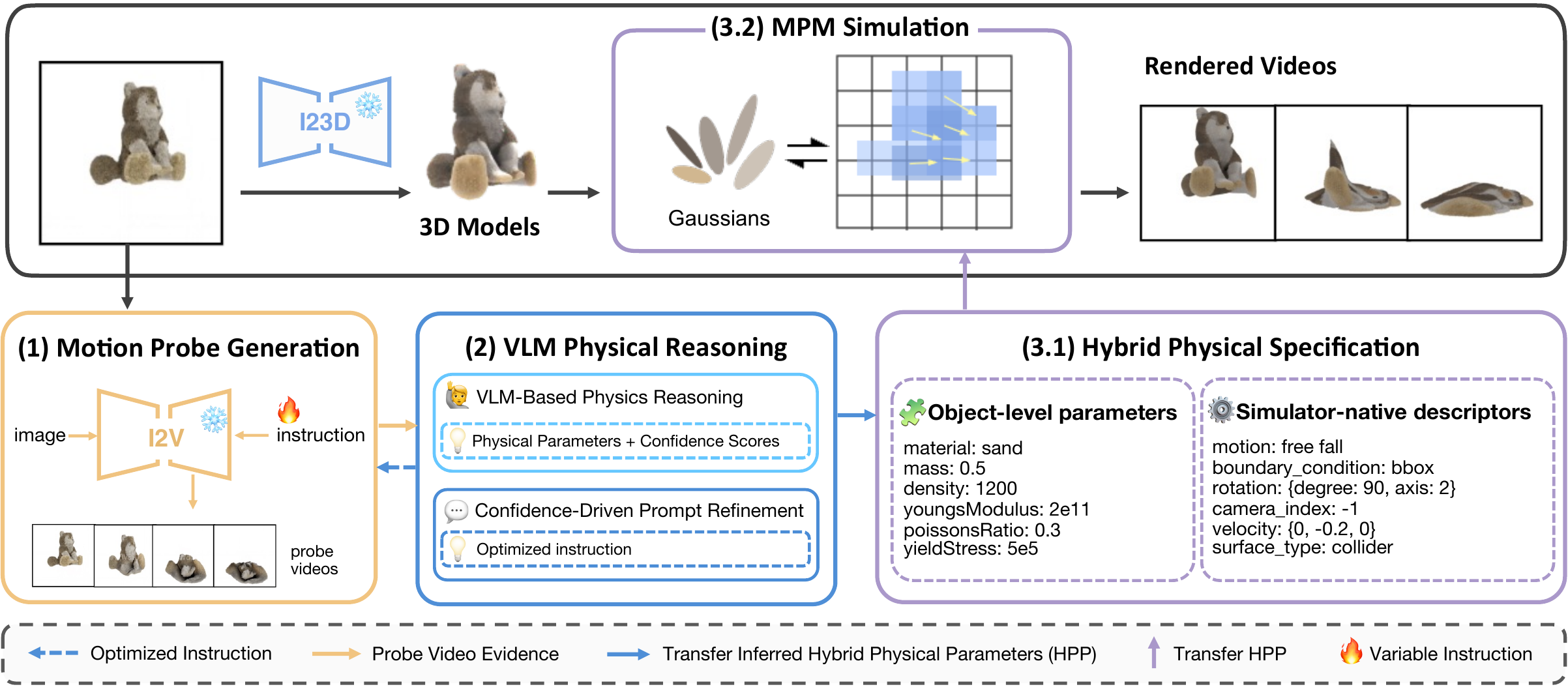}
    \caption{
    \textbf{Overview of PhyMAGIC framework.}
    Given a single input image, PhyMAGIC first generates motion probe videos using a pretrained I2V diffusion model, which exposes diverse temporal evidence for physical reasoning.
    A VLM then estimates object-level physical parameters together with confidence scores.
    Low-confidence attributes trigger a prompt refinement loop that synthesizes additional motion probes to gather complementary evidence.
    Once parameters reach sufficient confidence, PhyMAGIC compiles a hybrid physical specification combining material properties with simulator-native descriptors.
    The resulting configuration is executed in a differentiable MPM simulator to produce physically consistent dynamic 3D behaviors.
    }
    \label{fig:pipeline}
\end{figure}

We propose \textbf{PhyMAGIC}, a training-free framework for inferring physical properties and simulating physically consistent 3D dynamics from a single image.
As illustrated in \autoref{fig:pipeline}, PhyMAGIC consists of three components:
(1) \emph{Motion probe generation} (\S\ref{sec:4.1}) synthesizes probe videos from the input image using a pretrained image-to-video diffusion model, providing temporal evidence that reveals otherwise unobservable physical cues.
(2) \emph{VLM-based physical reasoning} (\S\ref{sec:4.2}) performs object-level physical inference within a confidence-driven refinement loop, where the VLM and I2V model alternate to progressively improve parameter estimates.
(3) \emph{Physics-grounded 3D dynamics} (\S\ref{sec:4.3}) compiles the inferred attributes into a \emph{Hybrid Physical Parameters (HPP)} specification that combines material properties with simulator-native descriptors, and executes it in a differentiable MPM simulator.
The closed loop between motion probe generation and VLM inference iteratively reduces uncertainty in physical reasoning, while the simulation stage translates the converged estimates into physically consistent dynamic 3D content.

\subsection{Motion Probe Generation}
\label{sec:4.1}
Physical attributes such as density, yield stress, and external forces are only revealed through temporal behavior~\cite{louis2022learning_acmmm2022}.
To enrich the visual cues available from a single static image, we use a pretrained image-to-video (I2V) diffusion model as a motion probe generator.
Specifically, we employ CogVideoX-I2V\footnote{We select CogVideoX-I2V for its open-source availability and support for reproducible inference with fixed seeds.}~\cite{yang2024cogvideox_iclr2025} to synthesize a sequence of frames $V_0 = \{I_0, I_1, \dots, I_T\}$ from the input image $I_0$.
To reduce redundancy while preserving informative transitions, we apply motion-aware subsampling based on optical flow magnitude, retaining frames with significant motion changes.
Rather than treating the generated video as a ground-truth simulation, we leverage it as an informative probe that transforms latent physical priors into explicit motion cues, such as temporal trajectories and deformation patterns, providing the subsequent VLM reasoning module with richer evidence for physical property estimation.

\subsection{VLM-Based Physical Reasoning}
\label{sec:4.2}

\begin{figure}[t]
    \captionsetup{font=small}
    \centering
    \includegraphics[width=1.0\linewidth]{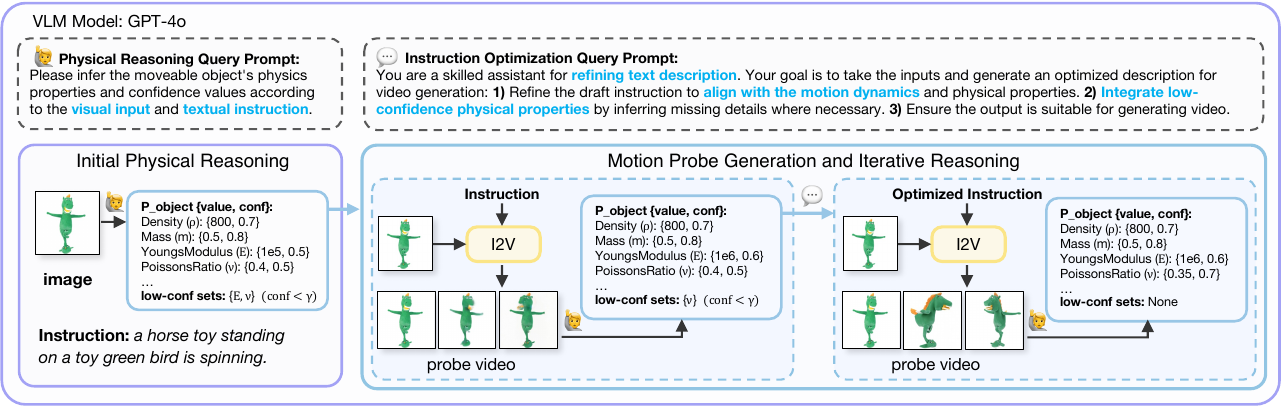}
    \caption{
    \textbf{Confidence-driven prompt refinement process.}
    Given an image and an initial instruction, a VLM (GPT-4o~\cite{achiam2023gpt_arxiv2023}) first infers coarse object-level physical properties together with confidence scores.
    Parameters with confidence below a threshold $\gamma$ are treated as uncertain and trigger motion probe generation.
    The probe videos provide complementary visual evidence, based on which the VLM refines the instruction to expose motion dynamics relevant to the uncertain properties.
    This process iteratively accumulates evidence until all parameters reach sufficient confidence or the maximum number of refinement iterations is reached, yielding a stable set of physical parameters for subsequent simulation.
    }
    \label{fig:confidence_driven_text_refinement}
\end{figure}

\noindent\textbf{Iterative Physical Inference.}
We design a structured, coarse-to-fine inference framework using GPT-4o~\cite{achiam2023gpt_arxiv2023}, which combines synthesized probe videos with targeted descriptive prompts.
As illustrated in \autoref{fig:confidence_driven_text_refinement}, the VLM first analyzes the input image and textual instruction to estimate a set of object-level physical parameters $P_i$ along with confidence scores $c_i \in [0, 1]$.
We estimate confidence via self-consistency: each attribute is inferred across up to three independent VLM samples, and $c_i$ is computed as the agreement ratio among the sampled responses.
Parameters with $c_i$ below a confidence threshold $\gamma$ are identified as uncertain and trigger the refinement mechanism described below.

\noindent\textbf{Confidence-Driven Prompt Refinement.}
Direct VLM predictions can suffer from prompt-motion misalignment or insufficient motion cues, leading to inaccurate estimates for certain attributes.
As illustrated in \autoref{fig:confidence_driven_text_refinement}, a toy horse is inferred with incorrect Young's modulus and Poisson's ratio due to vague descriptions and insufficient motion evidence.
To address this, we introduce a confidence-guided refinement loop.
We define a confidence threshold $\gamma$\footnote{
We select $\gamma{=}0.6$ using a held-out pilot set of 15 prompts (around 100 attributes queries). We plot the confidence-score histogram and choose $\gamma{=}0.6$ to separate the low-confidence sets that most often trigger video probe generation requests. See supplementary for the details.
% We set $\gamma{=}0.6$ based on empirical analysis: GPT-4o typically produces confidence scores around 0.7 for well-supported attributes, so $\gamma{=}0.6$ effectively isolates the few parameters that require additional evidence.
} and identify attributes $P_i$ whose scores fall below it.
These attributes are marked using a binary indicator:
\begin{equation}
    m_i =
    \begin{cases}
    1, & c_i < \gamma,\\
    0, & c_i \ge \gamma,
    \end{cases}
\end{equation}
where all indices $i$ with $m_i=1$ form the set requiring refinement.
Starting from the initial prompt $p^{(0)}$, we update it iteratively at step $t$ as
\begin{equation}
    p^{(t+1)} = \mathbf{VLM}(p^{(t)}, \{P_i \mid m_i=1\}),
\end{equation}
where the VLM receives the current prompt and low-confidence attributes, then generates a refined prompt that explicitly requests more informative motion patterns for those attributes.
At each iteration, the refined prompt drives a new round of I2V probe generation, and the VLM re-estimates all parameters with updated confidence scores.
This loop repeats until all attributes exceed $\gamma$ or a maximum iteration budget is reached.

\subsection{Physics-Grounded 3D Dynamics}
\label{sec:4.3}
While \S\ref{sec:4.2} focuses on identifying \emph{what} physical properties an object has, this stage addresses \emph{how} it moves and interacts over time.
We combine the inferred attributes with simulator-native descriptors into a unified specification, reconstruct a 3D representation from the input image, and execute the resulting configuration in a differentiable MPM solver.

\noindent\textbf{Hybrid Physical Parameters (HPP).}
A key challenge in connecting VLM reasoning to physical simulation is the semantic gap between their outputs.
The VLM produces high-level material descriptions (e.g., ``rigid body with density 1200 kg/m$^3$''), while the MPM solver requires a complete, low-level executable configuration specifying not only material constants but also boundary conditions, contact modes, and applied forces.
Neither component alone is sufficient for physically consistent simulation.
Material properties such as density and elasticity define how an object responds to forces, but without specifying the force itself, the initial velocity, or the boundary geometry, the simulator has no well-posed initial-boundary value problem to solve.
Conversely, execution context such as ``apply gravity and drop'' is meaningless without knowing whether the object is made of rubber or steel, as the same force profile produces qualitatively different trajectories depending on material stiffness and yield behavior.

To bridge this gap, we introduce a Hybrid Physical Parameters (HPP) specification that unifies both into a single simulator-ready configuration.
Formally, we define HPP as:
\begin{equation}
    \mathcal{H} = (\mathcal{P}_{\text{mat}},\; \mathcal{D}_{\text{sim}}),
\end{equation}
where $\mathcal{P}_{\text{mat}} = \{\rho, E, \nu, \sigma_y, \ldots\}$ denotes the object-level material properties (density, Young's modulus, Poisson's ratio, yield stress, etc.) inferred during the perception stage, and $\mathcal{D}_{\text{sim}} = \{v_0, \mathcal{B}, f_{\text{ext}}, \ldots\}$ denotes the simulator-native descriptors (initial velocity, boundary conditions, external forces, etc.).
All material values in $\mathcal{P}_{\text{mat}}$ are expressed in SI units~\cite{thompson2008use} and normalized into simulator ranges to maintain numerical stability.
HPP is defined at the object level and propagated to all constituent particles during simulation initialization.

For the simulator-native descriptors $\mathcal{D}_{\text{sim}}$, the VLM analyzes the textual instruction and accumulated probe evidence to select a canonical motion template from a predefined set (e.g., free-fall, compression, spinning, rebound, side-push).
Each template defines a structured combination of boundary type, initial velocity direction, force magnitude, and contact configuration.
For example, a ``side push'' template configures a horizontal initial velocity, a bounding-box boundary condition, and a collider surface type, while a ``free fall'' template applies only gravitational force with no initial velocity.
This template-based design ensures that the inferred physical state is translated into a complete, solver-compatible configuration while keeping the mapping interpretable and reproducible.

\noindent\textbf{MPM-Based 3D Dynamic Simulation.}
We reconstruct the input image into a 3D Gaussian Splatting representation using Trellis~\cite{xiang2024trellis_arxiv2024}.
Each Gaussian kernel serves as a material-carrying particle, initialized by augmenting the Gaussian state with the inferred HPP attributes:
\begin{equation}
    \mathcal{G}_p^t = (x_p^t, \Sigma_p^t, \alpha_p, c_p, \theta_p^t),
\end{equation}
where $\theta_p^t$ encodes physical properties such as density, elasticity, and yield stress.
We execute the simulation using a differentiable MPM solver~\cite{jiang2015affine_tog2015}, which updates particle positions, velocities, and deformation gradients via standard particle-grid-particle transfers.
Unlike existing differentiable simulators~\cite{abou2022particlenerf, le2023differentiable_ral2023, zhong2024reconstruction_eccv2024} that assume predefined material parameters, our approach initializes all physical attributes directly from VLM inference, enabling physically consistent dynamic 3D generation without manual parameter specification.

\section{Experiments}
\label{sec:experiments}
We evaluate PhyMAGIC in a training-free setting by integrating pretrained Trellis~\cite{xiang2024trellis_arxiv2024} for 3D Gaussian reconstruction, CogVideoX-5b~\cite{yang2024cogvideox_iclr2025} for video generation (720p, 50 frames), GPT-4o~\cite{achiam2023gpt_arxiv2023} for physical reasoning, and a differentiable MPM solver~\cite{xie2024physgaussian_cvpr2024} implemented in Warp~\cite{macklin2022warp_gtc2022}. 
Simulations utilize 128–200 steps, depending on scenario complexity, and run on a single NVIDIA RTX 4090 GPU. 
We benchmark on PhysGaussian~\cite{xie2024physgaussian_cvpr2024}, PhysGen~\cite{liu2024physgen_eccv2024}, and Internet-collected single-image scenes. 
Evaluation metrics include CLIP similarity for text–motion alignment, aesthetic score for visual quality, Image-Motion-FID for evaluating the distribution of ground-truth images with generated videos, and a 61-participant user study for physical plausibility and text consistency. 
We compare with state-of-the-art video generation models (CogVideoX, Open-Sora 2.0) and physics-aware approaches (PhysDreamer, Physics3D, OMNIPHYSGS). 
CogVideoX* denotes CogVideoX augmented with our confidence-driven prompt refinement for ablation analysis. 
Additional implementation settings, video generator choice, and user study details are provided in the supplementary material.

\begin{table*}[t]
\centering
\footnotesize
\captionsetup{font=small}
\setlength{\abovecaptionskip}{1pt}
\setlength{\belowcaptionskip}{4pt}
\caption{Quantitative evaluation on CLIP similarity and Aesthetic score with video generation methods across eight scenarios. Best results in each scenario are in \textbf{bold}, and the second-best results are in \underline{underline}.}
\label{table:csim_ascore_videos}

% ---------- CLIP similarity ----------
{\scriptsize \textbf{CLIP similarity} $\uparrow$}

\resizebox{0.98\textwidth}{!}{%
\renewcommand{\arraystretch}{1.2}
\setlength{\tabcolsep}{4pt}
\begin{tabular}{@{}l *{8}{c} >{\columncolor{green!8}}c@{}}
\toprule
\textbf{Method}
& \textbf{\makecell[c]{Swing \\ ficus}}
& \textbf{\makecell[c]{Sand \\wolf}}
& \textbf{\makecell[c]{Driving \\car}}
& \textbf{\makecell[c]{Rolling \\basketball}}
& \textbf{\makecell[c]{Tear \\toast}}
& \textbf{\makecell[c]{Sway \\tree}}
& \textbf{\makecell[c]{Lifting \\hat}}
& \textbf{\makecell[c]{Swing \\carnation}}
& \textbf{Average} \\
\midrule

OpenSora2.0
& 0.270 & \textbf{0.264} & 0.195 & 0.230 & 0.274 & 0.174 & 0.239 & 0.217 & \cellcolor{green!8}0.233 \\

CogVideoX
& 0.279 & 0.225 & 0.241 & 0.226 & 0.250 & 0.186 & 0.252 & 0.254 & \cellcolor{green!8}0.239 \\

CogVideoX*
& 0.257 & 0.263 & 0.234 & 0.233 & \textbf{0.279} & 0.195 & 0.241 & 0.254 & \cellcolor{green!8}\underline{0.240} \\

\rowcolor{gray!8}
\textbf{Ours}
& \textbf{0.294} & 0.200 & \textbf{0.252} & \textbf{0.266} & 0.231 & \textbf{0.243} & \textbf{0.270} & \textbf{0.256} & \cellcolor{green!8}\textbf{0.251} \\

\bottomrule
\end{tabular}
}

% ---------- Aesthetic score ----------
{\scriptsize \textbf{Aesthetic score} $\uparrow$}

\resizebox{0.98\textwidth}{!}{%
\renewcommand{\arraystretch}{1.2}
\setlength{\tabcolsep}{4pt}
\begin{tabular}{@{}l *{8}{c} >{\columncolor{green!8}}c@{}}
\toprule
\textbf{Method}
& \textbf{\makecell[c]{Swing \\ ficus}}
& \textbf{\makecell[c]{Sand \\wolf}}
& \textbf{\makecell[c]{Driving \\car}}
& \textbf{\makecell[c]{Rolling \\basketball}}
& \textbf{\makecell[c]{Tear \\toast}}
& \textbf{\makecell[c]{Sway \\tree}}
& \textbf{\makecell[c]{Lifting \\hat}}
& \textbf{\makecell[c]{Swing \\carnation}}
& \textbf{Average} \\
\midrule

OpenSora2.0
& 27.26 & 14.58 & 5.51 & 17.74 & 23.85 & 17.44 & 2.53 & 26.97 & 16.98 \\

CogVideoX
& 33.19 & 22.55 & 38.55 & 11.91 & 47.18 & 21.51 & 16.37 & \textbf{54.02} & 30.66 \\

CogVideoX*
& \textbf{34.89} & \textbf{23.94} & \textbf{41.20} & 17.95 & 46.49 & 22.53 & 16.41 & 51.47 & \textbf{31.86} \\

\rowcolor{gray!8}
\textbf{Ours}
& 31.00 & 18.38 & 29.40 & \textbf{26.82} & \textbf{49.85} & \textbf{39.67} & \textbf{26.74} & 23.70 & \cellcolor{green!8}\underline{30.69} \\

\bottomrule
\end{tabular}
}
\end{table*}

\begin{table*}[t]
  \centering
  \footnotesize
  \setlength{\abovecaptionskip}{1pt}
  \setlength{\belowcaptionskip}{2pt}
  \captionsetup{font=small}
  \caption{Quantitative results on CLIP similarity with physics-aware generation methods across six representative scenes. Best results are highlighted in \textbf{bold}.}
  \label{table:csim_physics}
  \scalebox{0.80}{
    \renewcommand{\arraystretch}{1.12}%
    \setlength{\tabcolsep}{8pt}%
    \begin{tabular}{@{}l c *{5}{c} >{\columncolor{green!8}}c@{}}
      \toprule
      % \textbf{Method} & \textbf{Swing ficus} & \textbf{Sand wolf} & \textbf{Lifting hat} & \textbf{Sway tree} & \textbf{Swing carnation} & \textbf{Rolling basketball} & \textbf{Average} \\
      \textbf{Method} & \textbf{\makecell[c]{Swing \\ ficus}} & \textbf{\makecell[c]{Sand \\ wolf}} & \textbf{\makecell[c]{Lifting \\ hat}} & \textbf{\makecell[c]{Sway \\ tree}} & \textbf{\makecell[c]{Swing \\ carnation}} & \textbf{\makecell[c]{Rolling \\ basketball}} & \textbf{Average} \\
      \midrule
      OMNIPHYSGS   & 0.227 & 0.167 & - & - & - & - & 0.197 \\
      \addlinespace[0.1em]
      PhysDreamer  & 0.223 & 0.145 & 0.239 & 0.186 & \textbf{0.272} & - & 0.213 \\
      \addlinespace[0.1em]
      Physics3D    & 0.225 & 0.147 & 0.229 & 0.147 & 0.269 &\textbf{0.283} & 0.217 \\
      \addlinespace[0.1em]   
      \rowcolor{gray!10} \textbf{Ours} & \textbf{0.294} & \textbf{0.200} & \textbf{0.270} & \textbf{0.226} & 0.256 & 0.266 & \cellcolor{green!8}\textbf{0.252} \\
      \bottomrule        
    \end{tabular}
  }
\end{table*}

\begin{table*}[t] 
  \begin{minipage}[t]{0.36\textwidth}
    \centering
    \footnotesize
    \setlength{\abovecaptionskip}{1pt}
    \setlength{\belowcaptionskip}{2pt}
    \renewcommand{\arraystretch}{1.12}
    \captionsetup{font=small}
    \caption{Quantitative results on Image and Motion-FID of generated videos with the GT input images. }
    \label{table:image_motion_fid}
    \scalebox{0.69}{
    \renewcommand\arraystretch{1.1}
    \setlength{\tabcolsep}{4pt}
    \begin{tabular}{@{}lc@{}}
      \toprule
      \textbf{Method} & \textbf{Image-Motion-FID$\downarrow$}\\ 
      \midrule
      PhysDreamer & 107.47 \\ 
      \addlinespace[0.1em]
      OMNIPHYSGS & 106.60\\ 
      \addlinespace[0.1em]
      Physics3D & 98.91\\ 
      \addlinespace[0.1em]
      \rowcolor{gray!10} \textbf{Ours} & \textbf{94.69} \\ 
      \bottomrule        
    \end{tabular}
    }
  \end{minipage}
  \hfill
  \begin{minipage}[t]{0.61\textwidth}
    \centering
    \footnotesize
    \setlength{\abovecaptionskip}{1pt}
    \setlength{\belowcaptionskip}{2pt}
    \renewcommand{\arraystretch}{1.12}
    \captionsetup{font=small}
    \caption{Human evaluation on physical plausibility and text consistency across eight real-world scenes compared to three I2V generation models. Ours achieves the highest human scores in both terms.}
    \setlength{\abovecaptionskip}{1pt}
    \label{table:use_study}
    \scalebox{0.69}{
    \renewcommand\arraystretch{1.1}
    \setlength{\tabcolsep}{8pt}
    \begin{tabular}{@{}lcc@{}}
      \toprule
      \textbf{Method} & \textbf{Physical Plausibility$\uparrow$} & \textbf{Text Consistency$\uparrow$} \\ 
      \midrule
      OpenSora2.0 & 2.02 & 2.18 \\ 
      \addlinespace[0.1em]
      CogVideoX & 2.58 & 2.49 \\ 
      \addlinespace[0.1em]
      CogVideoX\textsuperscript{*} & 2.97 & 3.00 \\ 
      \addlinespace[0.1em]
      \rowcolor{gray!10} \textbf{Ours} & \textbf{3.00} & \textbf{3.07} \\ 
      \bottomrule        
    \end{tabular}
    }
  \end{minipage}
\end{table*}

\begin{table*}[t]
\centering
\setlength{\abovecaptionskip}{1pt}
\setlength{\belowcaptionskip}{2pt}
\captionsetup{font=small}
\caption{Representative motion scenarios including ficus swing (elastic), car driving (rigid-body), and basketball rolling (deformable). 
We report key governing parameters (density, modulus, and friction angle) and their accuracy across iterations, showing progressive improvement and convergence toward ground truth.}
\label{table:ablation_study}
\resizebox{\textwidth}{!}{
\renewcommand\arraystretch{1.12}
\setlength{\tabcolsep}{1.2mm}
\begin{tabular}{l *{3}{cccc}}
\toprule
& \multicolumn{4}{c}{\textbf{Swing ficus}} & \multicolumn{4}{c}{\textbf{Rolling basketball}} & \multicolumn{4}{c}{\textbf{Driving car}} \\
\cmidrule(lr){2-5} \cmidrule(lr){6-9} \cmidrule(lr){10-13}
{\textbf{Parameter}} 
& \cellcolor{gray!15}\textbf{GT} & \textbf{Iter1} & \textbf{Iter2} & \textbf{Iter3 (final)}
& \cellcolor{gray!15}\textbf{GT} & \textbf{Iter1} & \textbf{Iter2} & \textbf{Iter3 (final)}
& \cellcolor{gray!15}\textbf{GT} & \textbf{Iter1} & \textbf{Iter2} & \textbf{Iter3 (final)} \\
\midrule
Material          
& \cellcolor{gray!15}Elastic & Elastic & Elastic & Elastic
& \cellcolor{gray!15}Elastic & Elastic & Elastic & Elastic
& \cellcolor{gray!15}Rigid   & Elastic & Rigid   & Rigid \\
Density           
& \cellcolor{gray!15}400   & 600   & 250   & 250
& \cellcolor{gray!15}1000  & 85.71 & 200   & 600
& \cellcolor{gray!15}1200  & 1000  & 1200  & 1200 \\
Young's Modulus   
& \cellcolor{gray!15}3e6   & 5e4   & 3e6   & 3e6
& \cellcolor{gray!15}1e5   & 1e4   & 1e5   & 1e5
& \cellcolor{gray!15}2e9   & 2.5e4 & 2e9  & 2e9 \\
Poisson's Ratio   
& \cellcolor{gray!15}0.30   & 0.30   & 0.40   & 0.30
& \cellcolor{gray!15}0.40   & 0.47  & 0.50   & 0.40
& \cellcolor{gray!15}0.40  & 0.25  & 0.75  & 0.50 \\
Yield Stress      
& \cellcolor{gray!15}--    & --    & --    & -- 
& \cellcolor{gray!15}--    & --    & --    & -- 
& \cellcolor{gray!15}6e7   & --    & 3e6   & 3e7 \\ 
\midrule
\rowcolor{gray!10}
\textbf{Accuracy (\%) $\uparrow$} 
& \cellcolor{gray!15}-- & \textbf{62.92} & \textbf{82.29} & \textbf{90.63}
& \cellcolor{gray!15}-- & \textbf{50.27} & \textbf{73.75} & \textbf{90.00}
& \cellcolor{gray!15}-- & \textbf{29.17} & \textbf{44.50} & \textbf{93.75} \\
\bottomrule
\end{tabular}
}
\end{table*}

\subsection{Quantitative Results}
\textbf{Comparison with image-to-video generation models.}
By leveraging iterative VLM-guided reasoning and a physically grounded MPM simulator, PhyMAGIC generates physically plausible 3D dynamics from only a single input image and text prompt.
\autoref{table:csim_ascore_videos} reports quantitative comparisons with state-of-the-art image-to-video generation models.
Our method achieves the highest CLIP similarity scores in most scenarios, indicating stronger semantic alignment between generated motions and textual descriptions.
On average, PhyMAGIC reaches 0.251, outperforming OpenSora2.0 (0.233), CogVideoX (0.239), and CogVideoX\textsuperscript{*} (0.240).
Despite operating with only a single-image input, PhyMAGIC also achieves competitive performance on the Aesthetic score, surpassing other models in the \textit{rolling basketball}, \textit{tear toast}, \textit{sway tree}, and \textit{lifting hat} scenarios.
These improvements stem from PhyMAGIC’s explicit integration of physical priors, which guides motion synthesis toward trajectories that are not only visually appealing but also physically consistent.

\noindent\textbf{Comparison with physics-aware 3D dynamic generation models.}
\autoref{table:csim_physics} compares PhyMAGIC with three representative physics-aware baselines across multiple challenging scenarios. 
PhyMAGIC achieves substantial gains, attaining scores of 0.294 and 0.200 on the elastic (\textit{swing ficus}) and sand (\textit{sand wolf}) cases, respectively. 
We present the distribution of the collected GT images with the generated videos in \autoref{table:image_motion_fid}. Our method achieves the lowest Image-Motion-FID score, surpassing existing physics-aware baselines.
These results highlight PhyMAGIC’s ability to preserve visual fidelity while maintaining accurate text–motion coherence. 
Notably, PhyMAGIC improves semantic similarity by 16.1\% over Physics3D, underscoring the advantage of our iterative VLM-guided refinement in producing dynamics that are both physically plausible and semantically aligned.

\noindent\textbf{Human Evaluation.}
We conducted a user study to assess physical plausibility and text–motion alignment.
Participants rated 32 videos on a 4-point Likert scale (1 = least, 4 = most consistent).
As shown in \autoref{table:use_study}, PhyMAGIC achieved the highest mean ratings (3.00 for plausibility, 3.07 for alignment), followed by CogVideoX\textsuperscript{*} (2.97/3.01), which still outperformed CogVideoX and OpenSora2.0.
These results confirm that confidence-guided refinement improves dynamic realism, which enables PhyMAGIC to deliver motions preferred by human evaluators.

\begin{figure}[!t]
    \captionsetup{font=small}
    \centering
    \includegraphics[width=0.84\linewidth]{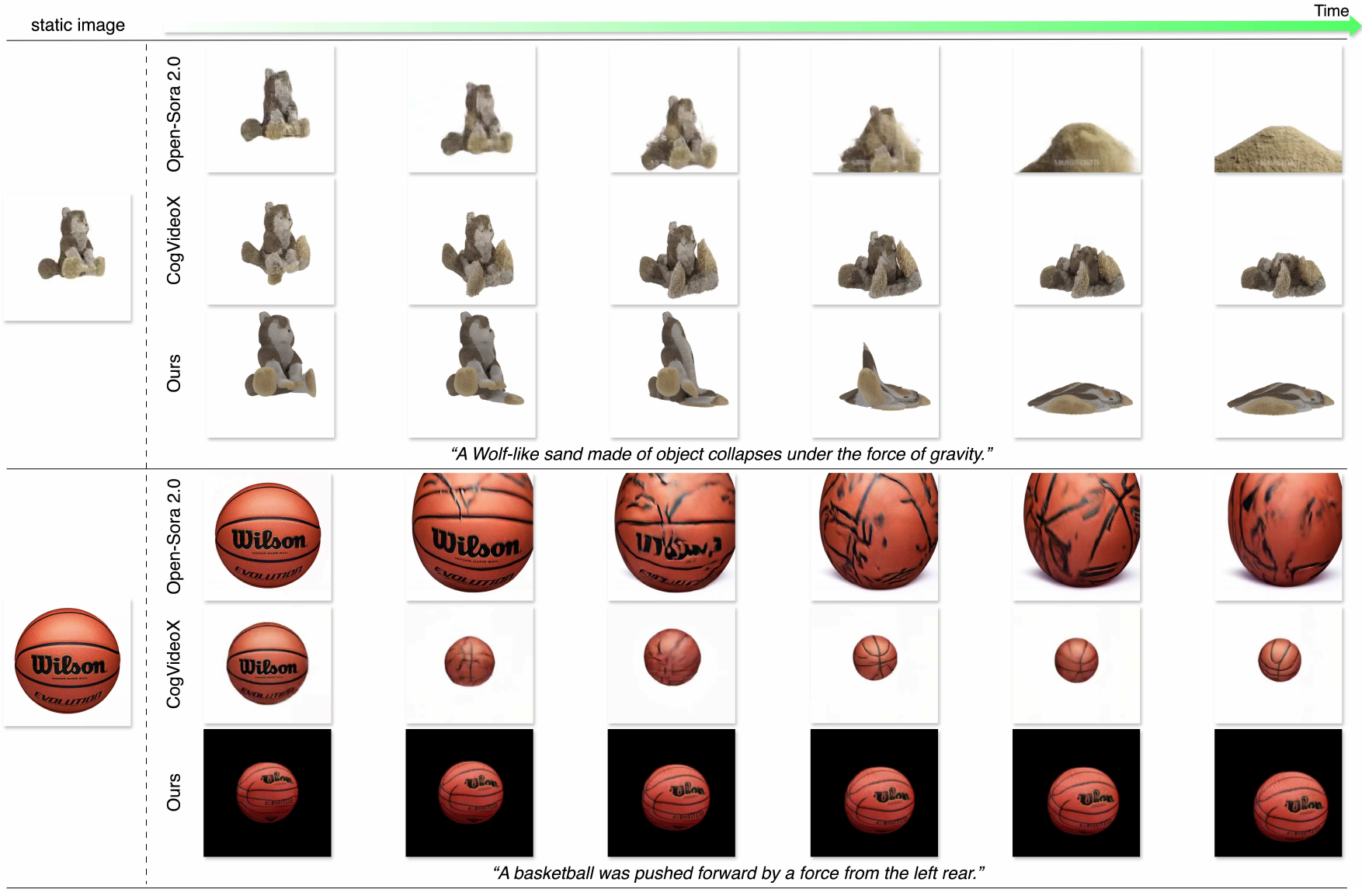}
    \caption{
    Qualitative comparisons of dynamic scene generation between our method and state-of-the-art video generation models~\cite{peng2025open_arxiv_2025, yang2024cogvideox_iclr2025}. 
    Given only a single static image, PhyMAGIC effectively infers intrinsic physical properties. It generates highly realistic dynamics over time, demonstrating superior physical realism and temporal consistency. 
    }
    \label{fig:qualitative_comparison}
\end{figure}
    
\begin{figure}[!t]
    \captionsetup{font=small}
    \centering
    \setlength{\abovecaptionskip}{3pt}
    \includegraphics[width=0.84\linewidth]{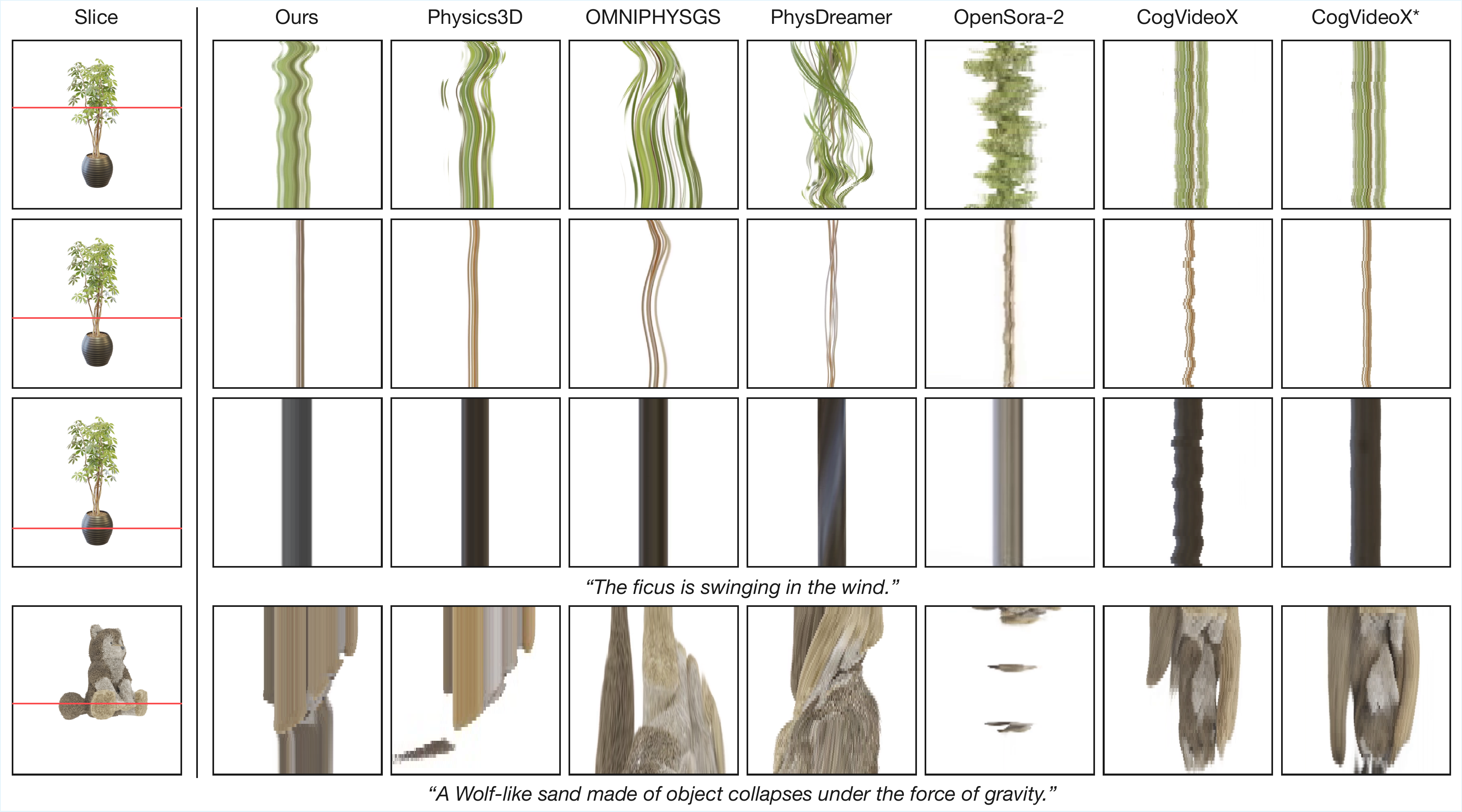}
    \caption{Space-time slices comparison with physical-aware and I2V generation methods.
}
\label{fig:slice_comparison}
\end{figure}

\subsection{Qualitative Results}

\textbf{Visual Comparison with Video Generation Models.}
As illustrated in \autoref{fig:qualitative_comparison}, our method exhibits clear advantages in modeling physically realistic dynamics given only a static image, whereas Open-Sora 2.0~\cite{peng2025open_arxiv_2025} and CogVideoX~\cite{yang2024cogvideox_iclr2025} often produce visual distortions or physically implausible motions. 
For the wolf-like sand collapse, Open-Sora 2.0 and CogVideoX exhibit unnatural particle scattering behaviors, whereas our results accurately simulate the behavior of sand material under gravity. 
In the \textit{rolling basketball} scene, baseline models produce implausible deformations under external forces or distorted appearance, whereas our method maintains consistent appearance and motion.

\noindent\textbf{Spatiotemporal Slice Analysis.}
To evaluate temporal coherence, we compare space–time slices in \autoref{fig:slice_comparison}. 
Our method produces precise and continuous motion trajectories that match expected physical laws. 
By contrast, Physics3D, OMNIPHYSGS, PhysDreamer, and OpenSora-2 often show blurred edges and fragmented strips, indicating weak temporal consistency and inaccurate physical dynamics.
CogVideoX and CogVideoX* exhibit regular motion patterns but fail to control static regions, resulting in unintended motion in non-moving components such as the base of the ficus.

\begin{figure}[t!]
    \captionsetup{font=small}
    \centering
    \includegraphics[width=0.96\linewidth]{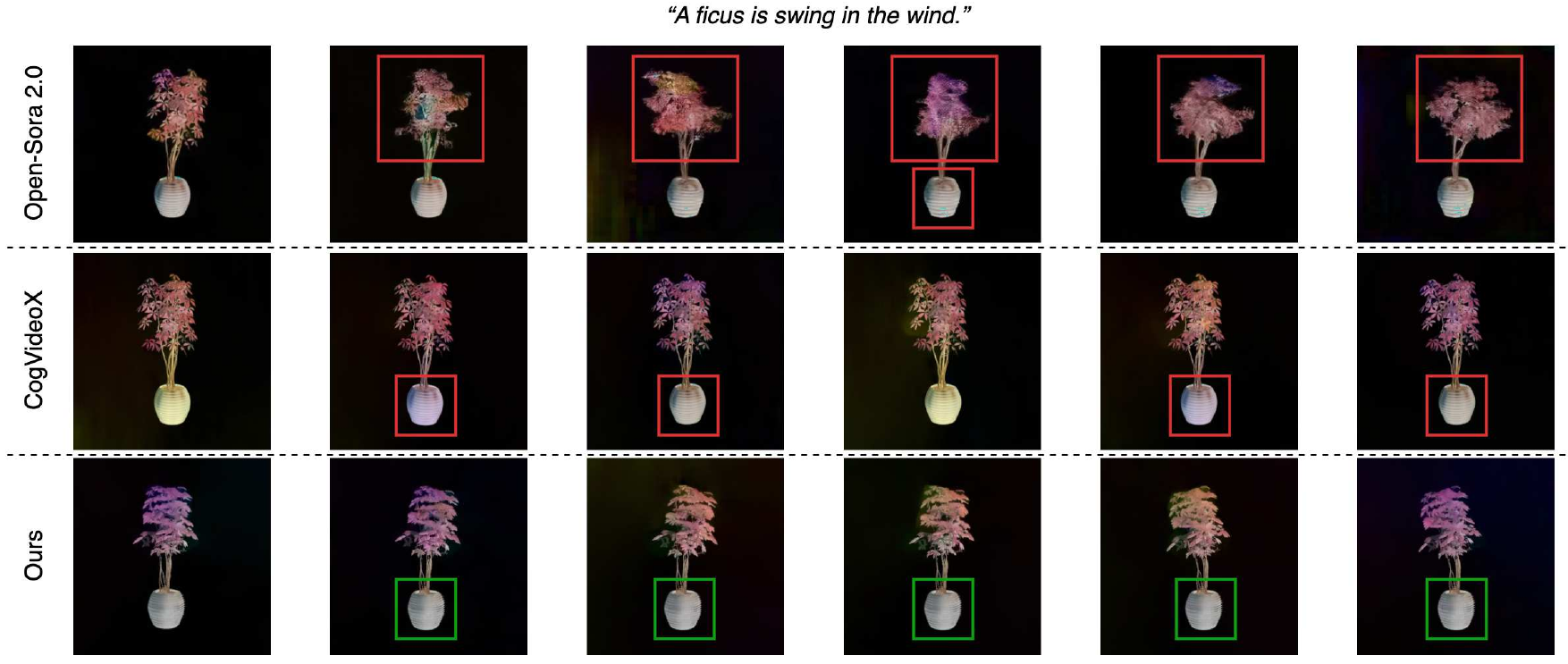}
\caption{
Qualitative comparison of optical flow visualizations on the \textit{swing ficus} scene.
PhyMAGIC maintains coherent motion and structural stability over time (green boxes), 
while competing methods~\cite{peng2025open_arxiv_2025, yang2024cogvideox_iclr2025} show structural inconsistencies and unnatural distortions (red boxes).
}
    \label{fig:optical_comparison}
\end{figure}

\noindent\textbf{Optical Flow Visualization.}
\autoref{fig:optical_comparison} visualizes optical flow~\cite{teed2020raft_eccv2020} on a sample \textit{swing ficus} scene.
OpenSora2.0 exhibits spurious edges and temporal jitter around movable regions, indicating weak motion coherence, and the bottom vase also shows slight deformation.
Both PhyMAGIC and CogVideoX illustrate stable optical flow; however, CogVideoX lacks precise spatial control and generates unintended motion in static regions due to its global motion modeling strategy.
Our method achieves accurate motion localization through physics-aware constraints, ensuring that only the target regions exhibit dynamic deformation while static objects remain unaffected.

\subsection{Ablation Study}
\label{subsec:ablation}
To assess the contribution of VLM-based physics reasoning and determine an effective number of refinement iterations, we report results in \autoref{table:ablation_study}.
With limited initial motion context, the first iteration often misclassifies materials (e.g., predicting a driving car as elastic) or yields inaccurate physical parameters.
Subsequent iterations progressively improve parameter estimation by refining text prompts and regenerating motions.
Performance stabilizes at three iterations, where the overall accuracy exceeds 90\%, indicating that iterative reasoning effectively resolves early ambiguities and converges to a reliable physical understanding.
More results are provided in the supplementary material.

\begin{figure}[t]
    \captionsetup{font=small}
    \centering
    \includegraphics[width=1.0\linewidth]{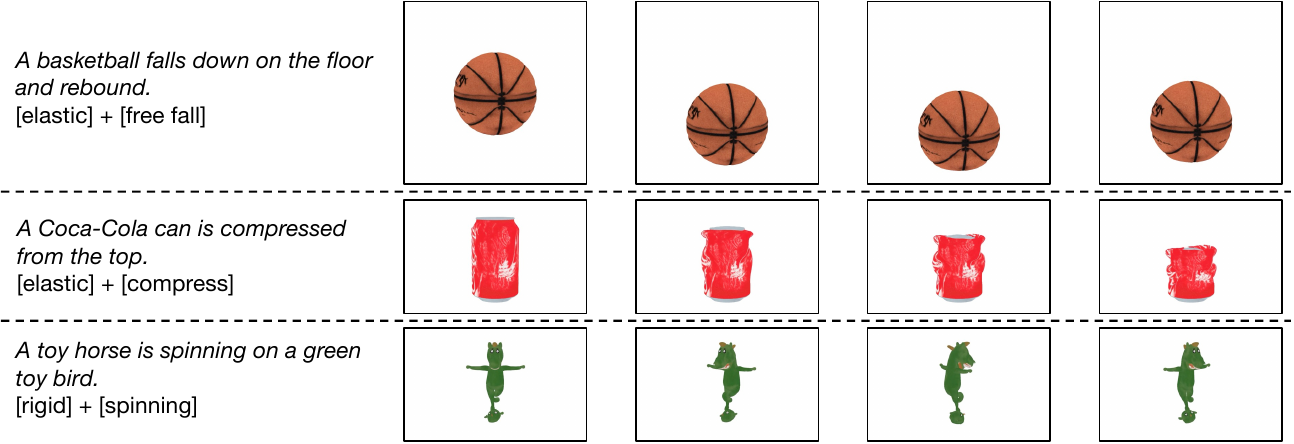}
    \caption{
    \textbf{The generalization ability of our PhyMAGIC across different materials and motion types.}
    Given a single image (the first column frame), PhyMAGIC produces plausible and controlled dynamics under diverse motion conditions.}
    \label{fig:generalization}
\end{figure}

% \begin{table*}[t]
% \centering
% \footnotesize
% \setlength{\tabcolsep}{2pt}
% \renewcommand\arraystretch{1.18}
% \begin{minipage}[t]{0.46\textwidth}
% \centering
% \caption{Iterative refinement ablation on the Material accuracy, BC accuracy, and Probe budget across five-scene.}
% \label{tab:iter_ablation}
% \resizebox{\linewidth}{!}{
% \begin{tabular}{lccc}
% \toprule
% Method & Mat. acc. (\%) $\uparrow$ & BC acc. (\%) $\uparrow$ & Probes/scene $\downarrow$ \\
% \midrule
% One-shot VLM + fixed boundary           & 60.0 & 60.0 & 0.00 \\
% One-shot VLM + best template            & 76.0 & 80.0 & 0.00 \\
% Iterative probing, no confidence gating & 83.0 & 80.0 & 1.00 \\
% PhyMAGIC, full                          & \textbf{90.0} & \textbf{90.0} & \textbf{0.80} \\
% \bottomrule
% \end{tabular}
% }
% \end{minipage}
% \hfill
% \begin{minipage}[t]{0.51\textwidth}
% \centering
% \caption{Comparison of OpenSora-2 and CogVideoX-5B as I2V models in our pipeline on accuracy gain over the single-image baseline, inference time, and GPU need.}
% \label{table:video_generator_choice}
% \resizebox{\linewidth}{!}{
% \begin{tabular}{lccc}
% \toprule
% \textbf{\makecell[c]{Video\\ Generators}} 
% & \textbf{\makecell[c]{Inference Accuracy\\Gain (±\%)}} 
% & \textbf{\makecell[c]{Inference\\Time (mins)}} 
% & \textbf{\makecell[c]{Max GPU Memory\\Required (GB)}} \\ 
% \midrule
% OpenSora-2    & +10.9 & 53 & $\sim$83 \\ 
% CogVideoX-5B  & +12.2 & 21 & $\sim$9 \\ 
% \bottomrule        
% \end{tabular}
% }
% \end{minipage}
% \end{table*}

\begin{table}[t]
\centering
\small
\setlength{\tabcolsep}{3pt}
\setlength{\abovecaptionskip}{1pt}
\setlength{\belowcaptionskip}{2pt}
\renewcommand{\arraystretch}{1.12}
\setlength{\tabcolsep}{6pt}
\caption{Iterative refinement ablation of material accuracy, boundary-condition accuracy, and the average number of probes per scene on the five-scene subset.}
\label{tab:iter_ablation}
\scalebox{0.82}{
\begin{tabular}{lccc}
\toprule
\textbf{Method} 
& \textbf{Mat. acc. (\%) $\uparrow$} 
& \textbf{BC acc. (\%) $\uparrow$} 
& \textbf{Probes/scene $\downarrow$} \\
\midrule
One-shot VLM + fixed boundary           
& 60.0 & 60.0 & 0.00 \\
One-shot VLM + best template            
& 76.0 & 80.0 & 0.00 \\
Iterative probing, no confidence gating 
& 83.0 & 80.0 & 1.00 \\
PhyMAGIC, full                          
& \textbf{90.0} & \textbf{90.0} & \textbf{0.80} \\
\bottomrule
\end{tabular}
}
\end{table}

\begin{table*}[t]
    \centering
    \footnotesize
    \setlength{\abovecaptionskip}{1pt}
    \setlength{\belowcaptionskip}{2pt}
    \captionsetup{font=small}
    \caption{Comparison of OpenSora-2 and CogVideoX-5B as I2V models in our pipeline on accuracy gain over the single-image baseline, inference time, and GPU need.}
    \label{table:video_generator_choice}
    \scalebox{0.86}{
    \renewcommand\arraystretch{1.22}
    \setlength{\tabcolsep}{9pt}
    \begin{tabular}{lccc}
      \toprule
      \textbf{\makecell[c]{Video\\ Generators}} & \textbf{\makecell[c]{Inference Accuracy \\Gain (±\%)}} & \textbf{\makecell[c]{Inference \\Time (mins)}} & \textbf{\makecell[c]{Max GPU Memory \\Required (GB)}}\\ 
      \midrule
      OpenSora-2	 & +10.9 & 53 & $\sim$83 \\ 
      \addlinespace[0.1em]
      CogVideoX-5B & +12.2 & 21 & $\sim$9 \\ 
      \bottomrule        
    \end{tabular}
    }
\end{table*}

\subsection{Analysis}
\noindent\textbf{Generalization.}
Due to the training-free and modular nature of PhyMAGIC, it exhibits robust generalization across distinct material-behavior combinations, ranging from elastic and rigid to granular material.
We showcase this diversity through a variety of motion patterns in~\autoref{fig:generalization}, including free-fall, collision, spinning, compression, and side-push. 

\noindent\textbf{Necessity of Iterative Probing.}
A one-shot VLM observes only a static image, while density, modulus, and yield stress are ambiguous. 
As shown in Table~\ref{tab:iter_ablation}, one-shot reaches 60.0\%/60.0\% material/BC accuracy, while PhyMAGIC-full reaches 90.0\%/90.0\% with 20\% less probe budget than ungated probing. 
This gain shows that iterative video probing provides missing dynamic evidence for physical inference, rather than simply increasing computation.
The confidence gate further avoids unnecessary probes when the initial prediction is already reliable, leading to better accuracy with a lower average probe cost.

\noindent\textbf{Video Generator Choice.}
Considering the limitation of computing resources, we experimented with both OpenSora-2~\cite{peng2025open_arxiv_2025} and CogVideoX~\cite{yang2024cogvideox_iclr2025} as candidate open-source image-to-video generators for our framework. 
We found that both models yielded comparable improvements in physical inference accuracy than single image, as shown in \autoref{table:video_generator_choice}, with a difference of less than 1.3\%. 
However, CogVideoX was substantially more resource-efficient, requiring only a single 9 GB GPU and 21 minutes for inference, whereas OpenSora-2 approximately needed 86 GB of GPU memory and 53 minutes of inference time. 
Based on this comparison, we selected CogVideoX as our primary video generation model due to its efficiency and open-source availability, which are key requirements for our iterative physical reasoning pipeline.

\noindent\textbf{Limitation.}
Despite its generalization, the simulation fidelity is inherently coupled with the quality of the initial 3D reconstruction.
For instance, thin-shell structures or complex topologies generated by TRELLIS~\cite{xiang2024trellis_arxiv2024} may result in suboptimal particle sampling for the MPM solver. 
Second, while our iterative motion-probing loop significantly reduces ambiguity, the current VLM-based reasoning is primarily qualitative. 
This may lead to discrepancies in high-precision scenarios, such as determining exact friction coefficients or modeling high-frequency elastic vibrations. 
Finally, the current MPM simulation focuses on foreground object dynamics and does not yet support complex articulated objects or dense multi-object interactions.
Future iterations could integrate differentiable rendering gradients to fine-tune these physical parameters directly from the visual residuals between the probe videos and the simulated results.
\section{Conclusions}
\label{sec:conclusion}
We present PhyMAGIC, a training-free framework that achieves motion-aware generation through iterative VLM-guided video generation and differentiable MPM simulation.  
Our innovations lie in integrating a pretrained video generation model and a confidence-driven VLM feedback mechanism for physics reasoning optimization in the iteration.
We also leverage a real-time MPM engine that simulates multi-material behaviors, guided by parameters inferred from robust physics reasoning. 
Extensive experimental results demonstrate that PhyMAGIC significantly improves physical plausibility in dynamic 3D assets, while maintaining high-quality text-video consistency.

\bibliographystyle{splncs04}
\bibliography{main}

\end{document}